\documentclass[11pt,a4paper]{article}

\usepackage[T1]{fontenc}
\usepackage[margin=1in]{geometry}
\usepackage{amsmath,amssymb}
\usepackage{amsthm}
\usepackage{booktabs}
\usepackage{graphicx}
\usepackage{placeins}
\usepackage[round,authoryear]{natbib}
\usepackage{xcolor}
\usepackage{newtxtext}
\usepackage[subscriptcorrection]{newtxmath}
\usepackage[
    unicode,
    hidelinks,
    bookmarksnumbered=true,
    bookmarksopen=true,
    bookmarksopenlevel=2,
    pdfpagemode=UseOutlines,
    hypertexnames=false
]{hyperref}
\usepackage{bookmark}
\hypersetup{
    pdftitle={Screening Collapse under Recursive Synthetic Data},
    pdfauthor={Jia Zhang and Xin Chen}
}

\newtheorem{theorem}{Theorem}

\newtheorem{corollary}{Corollary}
\newtheorem{proposition}{Proposition}
\theoremstyle{definition}
\newtheorem{definition}{Definition}

\newtheorem{remark}{Remark}

\newenvironment{keywords}{\par\smallskip\noindent\textbf{Keywords: }}{\par\smallskip}

\numberwithin{equation}{section}
\numberwithin{theorem}{section}
\numberwithin{lemma}{section}
\numberwithin{corollary}{section}
\numberwithin{proposition}{section}
\numberwithin{definition}{section}
\numberwithin{assumption}{section}
\numberwithin{remark}{section}
\numberwithin{step}{section}
\numberwithin{condition}{section}
\numberwithin{property}{section}
\numberwithin{restrictions}{section}
\numberwithin{example}{section}
\numberwithin{algo}{section}

\newcommand{\E}{\mathbb E}
\newcommand{\Pp}{\mathbb P}

\renewcommand{\paragraph}[1]{\par\addvspace{6pt}\noindent{\itshape #1}\quad}

\begin{document}

\title{Screening Collapse under Recursive Synthetic Data}
\author{Jia Zhang$^{1}$ \qquad Xin Chen$^{2,\ast}$\\[0.6em]
\small $^1$Joint Laboratory of Data Science and Business Intelligence,\\
\small Southwestern University of Finance and Economics, Chengdu 611130, China\\
\small \texttt{zhangjia@swufe.edu.cn}\\[0.4em]
\small $^2$Department of Statistics and Data Science,\\
\small Southern University of Science and Technology, Shenzhen 518055, China\\
\small \texttt{chenx8@sustech.edu.cn}\\[0.4em]
\small $^\ast$Corresponding author}
\date{}

\pdfbookmark[0]{Screening Collapse under Recursive Synthetic Data}{paper-root}
\maketitle

\pdfbookmark[1]{Abstract}{abstract-bookmark}
\begin{abstract}
Recursive synthetic-data training can degrade a fitted model, but its effects on the variable-screening stage that precedes model fitting are less understood. When each generation labels data for the next, a marginal screener ranks variables against the current pseudo-label target rather than the original response. Across generations, this target drift can erode or reverse the active--inactive score margin and cause screening to discard variables that clean-label screening would retain.
We call this process \emph{screening collapse}. In a Gaussian linear model with minimum-norm recursive generators, we establish finite-sample screening bounds, characterize a late-generation information-loss limit, and derive a minimax impossibility result. Screening collapse can precede and compound ordinary model collapse: omitted variables incur a covariance-weighted loss that no refit restricted to the selected variables can eliminate, while recursive labels further degrade estimation on the retained variables. We also examine two approaches to mitigating screening collapse. Fixed-fraction real-label retention preserves a clean-label component but leaves resolution fixed, whereas data accumulation can avoid screening collapse if the pooled target preserves a positive margin and the growing sample resolves it.
Numerical experiments on gene-expression data support the theoretical findings.
\end{abstract}

\begin{keywords}
Feature screening; High-dimensional prediction; Model collapse; Recursive synthetic data; Sure independence screening.
\end{keywords}

\section{Introduction}

Generative artificial intelligence has made model-produced text, images, and labels a potential source of training data for later systems. When these outputs are reused for training, errors in one generation become part of the data seen by the next. This feedback loop can progressively distort the original distribution and suppress tail information or diversity, a phenomenon known as model collapse \citep{shumailov2024collapse,alemohammad2024mad,seddik2024howbad}. Complementary regression and scaling-law analyses show that recursive synthetic data can increase predictive risk, degrade loss scaling, and induce skill unlearning \citep{dohmatob2024regression,dohmatob2024tails}. Separate accumulation theory shows that retaining earlier data can stabilize final-model risk \citep{gerstgrasser2024inevitable}. However, how recursive synthetic data affect variable screening before the final model is fitted remains less understood.

Variable screening is an important step in high-dimensional prediction, determining which variables remain available for downstream model fitting. A marginal screener reduces a large candidate set to a tractable subset, after which ridge regression, the lasso, or another estimator is fit. Under suitable signal and regularity conditions, screening approaches can retain the full active set with high probability even when the ambient dimension grows exponentially with sample size \citep{fanlv2018sisreview}. However, when model-generated outputs supply the labels used for screening, recursion can alter the ranking itself. A relevant variable discarded at this stage is unavailable to any downstream estimator, regardless of how well that estimator is tuned.

We call this phenomenon \emph{screening collapse}: recursive pseudo-labels cause a screening rule to discard features that would be retained under the original clean-label task. This phenomenon is not tied to a particular screening method. For a broad class of supervised screening rules based on ranked relevance scores, the labels determine the population scores used to order the variables. Recursive pseudo-labeling moves these scores away from their clean-label counterparts, thereby reducing or reversing the separation between relevant features and their strongest competitors. When this margin becomes comparable to the estimation error in the empirical scores, the selected set can become unstable and relevant features may be omitted. Covariate dependence can amplify target drift by redistributing relevance across variables, while randomness in the recursive labeler introduces further perturbations. Screening collapse is therefore governed by whether the recursively evolving population-score margin, after generator-induced perturbations, remains large enough to be resolved by the finite screening sample.

This mechanism distinguishes screening collapse from ordinary model collapse while linking the two. The screening stage determines the set of variables available to the final model, whereas the fitting stage determines how well the model performs within that set. Once screening omits an active variable, even an oracle refit restricted to the selected variables incurs a covariance-weighted omitted-support loss. Recursive labels can therefore impair prediction through two distinct pathways: by restricting the available representation and by degrading estimation on the retained variables. This decomposition makes the loss attributable to screening explicit rather than subsuming it into a single measure of final prediction risk.

It also changes what data accumulation must accomplish. Final-model analyses suggest that retaining earlier data can stabilize prediction relative to replacement \citep{gerstgrasser2024inevitable}, but this conclusion does not automatically extend to a pipeline with variable screening. At the screening stage, data accumulation retains all earlier batches, so both the pooled target and the screening sample size evolve. Screening collapse can be avoided if the pooled target retains a positive active--inactive margin and the growing sample estimates the marginal scores accurately enough to resolve it. Increasing the sample size can reduce estimation error, but it cannot by itself reverse a population ranking that has been reversed by covariate dependence. To isolate target preservation from sample-size growth, we also study fixed-fraction real-label retention as a benchmark. This strategy keeps the screening sample size fixed while assigning real labels to a fixed fraction of observations. It therefore preserves a clean-label component in the screening target but gains no additional resolution over generations.

To formalize these arguments, we first establish a generic margin-versus-resolution principle for score-ranking screeners and then specialize it to sure independence screening (SIS) \citep{fanlv2008sis} in a Gaussian sparse linear model with a noisy minimum-norm recursive labeler. Under replacement, we first characterize the large-generation regime, showing how the centered signal loses scale and how screening approaches an information-loss baseline as the initialization is forgotten. We then derive finite-sample bounds at fixed generations, separating generational target drift, generator-induced fluctuation, and empirical screening error. Next, we isolate the irreducible omitted-support component of downstream prediction risk. Finally, we compare fixed-fraction retention with data accumulation, distinguishing preservation of the screening target from growth in statistical resolution.


Our main contributions are as follows:
\begin{itemize}
    \item We identify screening collapse as a distinct upstream consequence of recursive synthetic supervision and formulate a generic margin-versus-resolution principle for marginal screeners.
    \item For the Gaussian recursive-label experiment, we establish late-generation information-loss bounds for arbitrary measurable screening procedures and prove a minimax impossibility result. We then derive SIS-specific screening guarantees at fixed generations.
    \item We show how screening collapse compounds downstream model collapse, and prove that
screening collapse can itself impose a nearly worst-case floor on downstream prediction risk.
    \item We propose fixed-fraction real-label retention and data accumulation to mitigate or avoid
screening collapse, showing that retention can mitigate screening collapse, and accumulation
can avoid screening collapse under certain conditions.
\end{itemize}
We further investigate these mechanisms through high-dimensional simulations under various covariance structures and recursive pseudo-label experiments involving multiple screening rules on high-dimensional gene-expression data.

\subsection{Related Work}

\paragraph{Synthetic-data feedback and model collapse.}
Model outputs can re-enter future datasets as annotations or observations,
creating feedback loops that amplify model-induced distortions
\citep{taori2023feedback}. Recursive training can then precipitate model collapse
by eroding distributional tails and diversity
\citep{shumailov2024collapse,alemohammad2024mad,seddik2024howbad}, increasing
predictive risk \citep{dohmatob2024regression}, or degrading loss scaling and
inducing skill unlearning \citep{dohmatob2024tails}.
Collapse is not determined by recursion alone: stability depends on the quality
of the initial approximation, the retained proportion of real data, and the
evolution of the training sample size
\citep{bertrand2024stability,fu2024selfconsuming}. In particular,
replacement and accumulation can have sharply different behavior
\citep{gerstgrasser2024inevitable,kazdan2025collapse}; even under accumulation,
however, avoiding collapse requires regularity conditions
\citep{barzilai2025iterative}. Closest to our setting,
\citet{wu2026selfdistillation} analyze repeated pseudo-label refitting in
overparameterized linear regression and show that it induces signal forgetting
and direction-dependent spectral filtering, while \citet{qiao2026selection}
show that biased sample selection can accelerate distributional collapse. The
former concerns final prediction risk and implicit spectral filtering, and the
latter selects observations rather than variables.
We instead study an explicit upstream variable screener and ask whether
recursively generated labels remove active covariates before the final model is
fit.

\paragraph{Ultrahigh-dimensional screening.}
SIS was introduced by \cite{fanlv2008sis} as a marginal-correlation filter for reducing the dimension of ultrahigh-dimensional linear models before second-stage estimation. Related screening methods cover generalized linear models, pseudo-likelihoods, sparse additive models, model-free dependence, categorical responses, distance-correlation screening and so on; see e.g., \cite{fan2009beyond}, \cite{fansong2010sisglm}, \cite{fanfengsong2011nis}, \cite{zhu2011modelfree}, \cite{li2012dcsis}, \cite{cui2015mvsis},  and \cite{mai2013kolmogorov}. Mutual-information screening follows the same dependence-ranking principle with estimators adapted to continuous or categorical labels \citep{kraskov2004estimating,ross2014mutual}. These works keep the population screening target fixed, while our target is generation-dependent because pseudo-label recursion changes the marginal signal.

\paragraph{Post-screening estimation and interpolation.}
Screening is often followed by SCAD, the Dantzig selector, the lasso, ridge regression, or another lower-dimensional estimator \citep{fanlv2008sis,fanli2001scad,candes2007dantzig,tibshirani1996lasso,hoerl1970ridge}. To distinguish screening loss from downstream fitting loss, we derive an estimator-agnostic risk decomposition: the first term is the screening-induced omitted-support loss, and the second is the refitting, shrinkage, and target error within the selected feature set.
In our theoretical setup,
the recursive label generator is a minimum-norm interpolator, linking the setup to ridgeless regression and benign overfitting \citep{belkin2019reconciling,bartlett2020benign,hastie2022ridgeless}. In the theoretical pipeline, interpolation generates the next round of supervision; screening determines what the eventual predictor is permitted to use.

\subsection{Paper Organization}
The remainder of the paper follows the pipeline from upstream screening to
downstream prediction and intervention. We begin in
Section~\ref{sec:screening-collapse} by motivating and defining screening
collapse and stating the generic margin principle. Section~\ref{sec:setup} then
develops the theoretical setup, including the sparse Gaussian linear model,
the recursive generator, and the generation-specific SIS problem. Building on
this setup, Section~\ref{sec:theory} analyzes replacement
recursion, first through its large-generation limit and then through
fixed-generation realized margins. Section~\ref{sec:prediction} then shows how
screening collapse compounds downstream model collapse by separating the
screening-induced omitted-support loss from downstream refitting error.
Mitigation is the focus
of Section~\ref{sec:accumulation}, where fixed-fraction real-label retention is
contrasted with data accumulation and its rank transition.
Section~\ref{sec:realdata-experiments} evaluates recursive screening and both
interventions on a public high-dimensional dataset; Section~\ref{sec:discussion}
then closes with implications and limitations. Proofs, auxiliary results,
Gaussian simulations, and the additional CELLxGENE/scTab experiment are
described in the Supplementary Material.

\section{What is screening collapse?}
\label{sec:screening-collapse}

Model collapse is usually studied at the end of the pipeline: a final model is
trained on recursively generated data and loses information about the original
distribution. High-dimensional pipelines have an earlier failure point. A
screening stage often chooses the variables before the final model is fit. If
that gate removes a relevant variable, later refitting can never use it.

Recursive pseudo-labels change the object being ranked. With clean labels, a
marginal screener ranks variables by their clean population dependence with the
response. With generation-\(g\) pseudo-labels, it ranks dependence with the
current recursive labeler. 
The selected set can drift before
any downstream model-collapse effect is measured.

\begin{figure}[ht]
    \centering
    \includegraphics[width=0.95\textwidth]{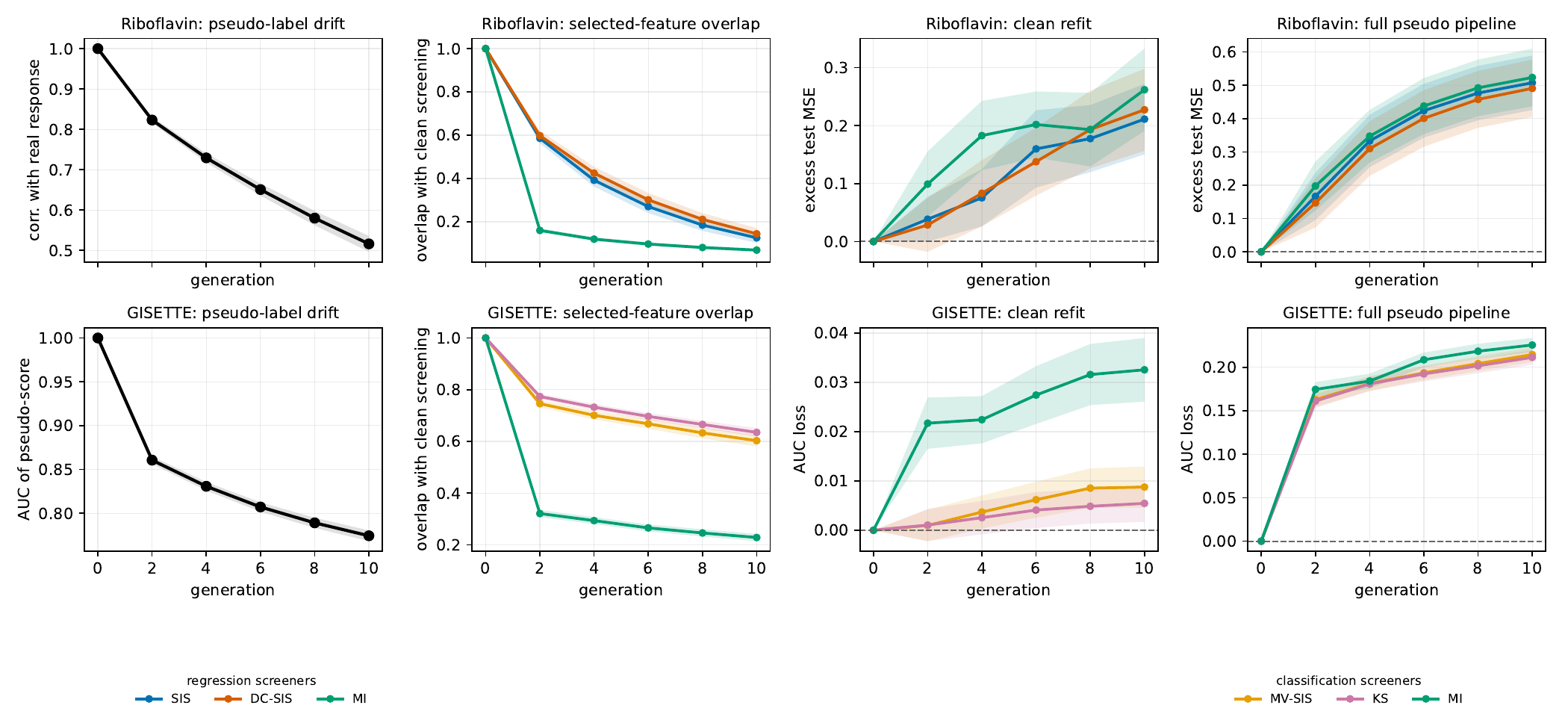}
    \caption[Recursive pseudo-labeling on Riboflavin]{Recursive pseudo-labeling on Riboflavin destabilizes three high-dimensional screening rules. 
    \protect\footnotemark\ Bands are pointwise 95\% normal-approximation intervals over 100 repeated train/test splits.}
    \label{fig:realdata-pseudolabel}
\end{figure}

Figure~\ref{fig:realdata-pseudolabel} shows the phenomenon in a public
high-dimensional regression dataset. Riboflavin contains 71 observations and
4088 gene-expression predictors \citep{buhlmann2014high}. A shallow random-feature
ridge smoother \citep{huang2006elm,rahimi2007randomfeatures} generates pseudo-labels across generations. This generator
allows us to test whether screening collapse persists under general nonlinear label
recursion, while avoiding the
training noise of a more complex neural network that could obscure the mechanism
under study. 
We compare SIS,
distance-correlation SIS, and mutual-information screening
\citep{fanlv2008sis,li2012dcsis,kraskov2004estimating}.
After screening, we fit an ordinary linear ridge model 
on the selected variables. The clean refit
uses real training labels, isolating selected-set drift; the full-pseudo fit uses
the current recursive labels and also captures downstream pseudo-label
degradation. Both are evaluated on held-out real labels.
Across these screeners, recursive pseudo-labeling reduces agreement with the
real labels, moves selected feature sets away from clean-label selections, and
increases the loss of clean refits. The common pattern is target drift, which we
formalize next.

\FloatBarrier
\footnotetext{Definitions of the metrics are given in Section~\ref{sec:realdata-experiments}.}

\begin{definition}[Recursive screening collapse]
\label{def:screening-collapse}
Recursive screening collapse is a generational process in which recursive
synthetic or pseudo-labels move the population target of an upstream screening
rule away from its clean-label target. As the target moves, screening loses the
variables or features that clean-label screening would retain. 
Early collapse
appears as erosion or sign reversal of the population screening margin. Late
collapse appears as convergence of the selected-set law to a baseline that no
longer carries the original signal.
\end{definition}

We use the full term when distinguishing this recursive mechanism from ordinary
clean-label screening failure, and write screening collapse when the recursive
setting is clear. The definition only needs a general generation-dependent score vector. For such scores, the following proposition formalizes how target drift can translate into variable loss.
Write \([p]=\{1,\ldots,p\}\). Fix a target set \(S_0\subset[p]\) with
\(1\le s=|S_0|\le p\), and fix a screening size \(D\) with \(s\le D\le p\). Let
\(T_g\in\mathbb R^p\) be the population scores of a marginal screening rule at
generation \(g\), and let \(\widehat T_g\) be their estimates from a screening
sample of size \(N\). For \(j\in[p]\), denote the corresponding coordinates by
\(T_{g,j}\) and \(\widehat T_{g,j}\), respectively. Let
\(\widehat S_g(D)\subset[p]\) be the indices of the
\(D\) largest values among \(\{|\widehat T_{g,j}|:j\in[p]\}\), with ties broken
by a fixed deterministic rule. For any score vector \(v\in\mathbb R^p\), define
\begin{equation}
    \tau(v,D)
    =
    \min_{j\in S_0}|v_j|
    -
    q_D\bigl(\{|v_k|:k\notin S_0\}\bigr),
    \label{eq:screening-margin}
\end{equation}
where \(q_D\) is the \((D-s+1)\)-st largest inactive magnitude, with value zero
if \(D-s+1>p-s\). This margin is the common bookkeeping device behind the
ranked-score screeners.

\begin{proposition}[Generic screening-margin principle]
\label{prop:generic-screening-principle}
Let \(\mathcal H_g\) be a sigma-field such that \(T_g\) is
\(\mathcal H_g\)-measurable. Let
\(\eta_{N,p}:(0,\infty)\to[0,\infty)\) be a deterministic error-bound function.
Suppose that, for some \(\varepsilon>0\),
\begin{align}
    \Pp\!\left\{
    \|\widehat T_g-T_g\|_\infty>\varepsilon
    \mid \mathcal H_g
    \right\}
    \le
    \eta_{N,p}(\varepsilon)
    \qquad\text{almost surely}. \label{eq:generic-screening-concentration}
\end{align}
Then, almost surely on \(\{\tau(T_g,D)>2\varepsilon\}\),
\[
    \Pp\{S_0\subset\widehat S_g(D)\mid\mathcal H_g\}
    \ge
    1-\eta_{N,p}(\varepsilon),
\]
and almost surely on \(\{\tau(T_g,D)<-2\varepsilon\}\),
\[
    \Pp\{S_0\subset\widehat S_g(D)\mid\mathcal H_g\}
    \le
    \eta_{N,p}(\varepsilon).
\]
\end{proposition}

Proposition~\ref{prop:generic-screening-principle} separates two sources of
randomness. The recursive mechanism fixes the realized population target
\(T_g\), possibly at random; conditional on that target, the fresh screening
sample only has to estimate \(T_g\) uniformly well. If the realized
active-inactive margin is positive and larger than the estimation error, sure
screening holds with high conditional probability. If the margin is negative by
the same amount, sure screening is unlikely. The proposition is therefore
conditional by design: unconditional collapse probabilities require a second
step, namely controlling the law of \(\tau(T_g,D)\) across generations.
Recursive labels matter because they move \(T_g\), so the realized margin can
shrink or reverse before any downstream model is fit. We next make the
separation between target drift and screening-sample concentration concrete with a sparse Gaussian linear model, 
and the canonical marginal screener
SIS. 

\section{Theoretical Setup}
\label{sec:setup}

Gaussian linear models provide a common tractable framework for studying model collapse under recursive synthetic data \citep{dohmatob2024regression,gerstgrasser2024inevitable}. Within this framework, we introduce the clean data model, recursive label-generation mechanism, and screening setup.

\subsection{Clean Sparse Linear Model and Recursive Generator}
The clean data are generated from the sparse linear model
\[
    y=x^\top w_0+\epsilon,
\]
where \(w_0\in\mathbb R^p\) contains the
original signal, whose support is \(S_0=\{j:w_{0,j}\ne0\}\) with
\(|S_0|=s\ge1\). The noise \(\epsilon\) is mean zero and independent of
\(x\). The covariate vector \(x\in\mathbb R^p\) is centered Gaussian,
\(x\sim N(0,\Sigma)\), with \(\Sigma\succ0\). For simplicity, the predictors
\(x_1,\ldots,x_p\) are standardized to have unit variance, so
\(\Sigma_{jj}=1\) for every \(j\in[p]\). We further assume
\(0<\lambda_*\le\lambda_{\min}(\Sigma)\le\lambda_{\max}(\Sigma)\le\Lambda\),
where \(\lambda_*\) and \(\Lambda\) are positive constants.

 We follow recursive synthetic-data setups for model collapse
\citep{dohmatob2024regression,shumailov2024collapse}, where each generation's
labeler supplies the training labels for the next generation.
Specifically, the recursive generator starts from the clean signal  \(\theta_0=w_0\).
Let \(\{X_g\}_{g\ge0}\) be independent matrices in \(\mathbb R^{r\times p}\)
with iid \(N(0,\Sigma)\) rows, where \(1\le r<p\). Independently of all
designs, let \(\eta_g\stackrel{\mathrm{iid}}{\sim}
N(0,\sigma_{\mathrm{gen}}^2I_r)\). At generation \(g\), generate labels
\[
    Y_g=X_g\theta_g+\eta_g,
\]
and let the next labeler be the minimum-norm least-squares interpolator
\[
    \theta_{g+1}=X_g^\dagger Y_g
    =P_g\theta_g+\zeta_g,
\]
where $X_g^\dagger = X_g^\top(X_gX_g^\top)^{-1}$, $P_g=X_g^\dagger X_g$ and \(\zeta_g=X_g^\dagger\eta_g\). Since a Gaussian \(X_g\) has row rank
\(r\) with probability one, this is the Moore--Penrose solution of the
underdetermined system \citep{penrose1955generalized,hastie2022ridgeless}.
When \(\sigma_{\mathrm{gen}}=0\), \(P_g\) is the orthogonal projection onto the
row space of \(X_g\). When \(\sigma_{\mathrm{gen}}>0\), the inverse Gram matrix
creates the additional noise component \(\zeta_g\).
We refer to this construction as replacement recursion: generation \(g+1\) is
trained only on the current batch \((X_g,Y_g)\), while all earlier batches are
discarded.

\subsection{SIS at Generation \texorpdfstring{\(g\)}{g}}
A fresh screening sample is independent of the generator. Its pairs
\((Z_i,\xi_i)\) are iid, with \(Z_i\) and \(\xi_i\) mutually independent:
\[
    Y_i^{(g)}=Z_i^\top\theta_g+\xi_i,
    \qquad Z_i\sim N(0,\Sigma),
    \qquad
    \xi_i\sim N(0,\sigma_{\mathrm{sis}}^2),
    \quad i\in[N],
\]
where $\sigma_{\mathrm{sis}}>0$ is a constant. Let
\(Z\in\mathbb R^{N\times p}\) have \(i\)-th row \(Z_i^\top\), and write
\(\xi=(\xi_1,\ldots,\xi_N)^\top\) and
\(Y_g^{\mathrm{scr}}=(Y_1^{(g)},\ldots,Y_N^{(g)})^\top=Z\theta_g+\xi\).
The SIS statistic at generation \(g\) is
\[
    \widehat\omega_j^{(g)}=\frac1N\sum_{i=1}^N Z_{ij}Y_i^{(g)}.
\]
Fix a screening size \(D\) with \(s\le D\le p\). The selected set
\(\widehat S_g(D)\) contains the \(D\) largest values of
\(|\widehat\omega_j^{(g)}|\). 

Conditional on \(\theta_g\), the SIS statistic is unbiased for the current marginal signal,
\[
    \E[\widehat\omega^{(g)}\mid\theta_g]
    =\Sigma\theta_g .
\]
Thus SIS ranks the current marginal signal \(\Sigma\theta_g\), not the original signal \(\Sigma w_0\). 
However,  we are interested in the unconditional probability that SIS retains all active variables, \(\Pp\{S_0\subset\widehat S_g(D)\}\).
The evaluation is anchored to the original clean distribution, not to
the synthetic law that supplies the generation-\(g\) labels. Thus \(g=0\)
corresponds to ordinary clean-data screening, while for \(g\ge1\) the screener,
and any downstream predictor trained after it, receives labels produced by a
recursive process outside the downstream model's control.


\begin{remark}
Although we use SIS as the concrete screener, the central
late-generation results below are not tied to its statistic.
Only the explicit
margin and score-concentration calculations use the SIS statistic; its simple
marginal form makes the evolving screening target and its empirical error
explicit and yields the transparent fixed-generation guarantee. Besides,
when \(\Sigma=I_p\), clean-label SIS is population-optimal
within marginal-correlation screeners and gives the cleanest starting point for
the collapse analysis. Under recursive
pseudo-labels, we show that even this favorable marginal screener can collapse once recursive
labels move its population target.
\end{remark}

\section{Theory of Screening Collapse}
\label{sec:theory}


In this section, we first ask what remains after many
generations, and then fix \(g\) to
study how the realized target and the screening sample determine the sure-screening event. Throughout this section, the collapse regime of interest assumes that
the clean marginal problem is screenable, i.e.,
\(\tau(\Sigma w_0,D)>0\), to focus on the recursive-data effect.

\subsection{Large-Generation Limit}
\label{subsec:large-generation-limit}

Notice that clean-label
screening ranks the marginal signal \(\Sigma w_0\), while recursive synthetic labels replace
this target by the random vector \(\Sigma\theta_g\), whose deterministic center is \(m_g:=\E[\Sigma\theta_g]\). 
At a fixed generation, the population margin \(\tau(\Sigma\theta_g,D)\) measures whether the population ranking retains all active variables and how difficult that ranking is to recover from a finite screening sample.
Since \(\theta_g\) is random, we instead track the deterministic centered screening margin $\tau_g^{\mathrm{cen}}(D)=\tau(m_g,D)$. 
The following proposition bounds the centered margin and shows that it vanishes
as \(g\to\infty\).
Denote \(M_\Sigma=\E[P_g]\) and
\(\rho_\Sigma=\lambda_{\max}(M_\Sigma)\). 

\begin{proposition}[Vanishing centered screening margin]
\label{prop:centered-margin-vanishes}
For every \(g\ge0\),
\[
    |\tau_g^{\mathrm{cen}}(D)|
    \le
    \Lambda\rho_\Sigma^g\|w_0\|_2 ,
\]
where $\rho_\Sigma<1$.
Consequently, for fixed \(r\) and \(p\),
\[
\tau_g^{\mathrm{cen}}(D)\to0,~~\text{as}~~ g\to\infty.
\]
\end{proposition}

In the identity covariance case, $M_\Sigma= (r/p) I_p$
and therefore $m_g=(r/p)^gw_0$, which gives the explicit formulas $\tau_g^{\mathrm{cen}}(D)=(r/p)^g
    \min_{j\in S_0}|w_{0,j}|$.
Here \(\tau_g^{\mathrm{cen}}(D)\) decreases to zero from the positive side and never becomes negative at finite \(g\). 
Finite-sample failure instead comes from trying to resolve a shrinking positive margin, and the late behavior is governed by the limiting screening law stated below in Theorem~\ref{thm:random-selection-limit}. 
Under general covariance, unequal attenuation and covariance-induced leakage can reverse the active--inactive ranking, so \(\tau_g^{\mathrm{cen}}(D)\) need not vanish monotonically, and it can even become negative, as illustrated by the block-covariance experiment in the Supplementary Material.



We then ask what screening law remains when $g\to\infty$. To isolate that endpoint, introduce a zero-start generator with \(\widetilde\theta_0=0\) and \(\widetilde\theta_{g+1}=P_g\widetilde\theta_g+\zeta_g\). Let \(\widetilde Y_g^{\mathrm{scr}}\) be its fresh screening-label vector and \(\widetilde S_g(D)\) the corresponding SIS set. Define
\(
    \Pi_{\Sigma,D,g}^{\mathrm{gen}}(S_0)
    =
    \Pp\{S_0\subset\widetilde S_g(D)\}
\).

\begin{theorem}[Large-generation information-loss baseline]
\label{thm:random-selection-limit}
For every \(g\ge0\),
\[
    \left|
    \Pp\{S_0\subset\widehat S_g(D)\}
    -
    \Pi_{\Sigma,D,g}^{\mathrm{gen}}(S_0)
    \right|
    \le
    \frac{\sqrt{N\Lambda}}{2\sigma_{\mathrm{sis}}}
    \rho_\Sigma^{g/2}\|w_0\|_2 .
\]
Moreover, 
if \(p>r+1\), 
the zero-start recursion has a unique stationary law
\(\mu_\infty\) with finite second moment. Let
\(\widetilde\theta_\infty\sim\mu_\infty\), independently of a fresh screening
sample, and let \(\widetilde S_\infty(D)\) be the corresponding SIS set. Define
$
    \Pi_{\Sigma,D,\infty}^{\mathrm{gen}}(S_0)
    =
    \Pp\{S_0\subset\widetilde S_\infty(D)\}
$.
Then, for fixed \(N,r\) and $p$, 
\[
    \Pp\{S_0\subset\widehat S_g(D)\}
    \to
    \Pi_{\Sigma,D,\infty}^{\mathrm{gen}}(S_0),~~\text{as}~~ g\to\infty.
\]
\end{theorem}

Although stated for SIS, Theorem \ref{thm:random-selection-limit} does not rely on the SIS statistic: the same
comparison applies to any  measurable screening procedure based on the
screening sample. The data-processing step in the Supplementary Material
makes this extension explicit.
Notice that the law \(\mu_\infty\) is not residual clean signal. It is the stationary footprint of repeated random projection and additive generator noise from a zero start. 
Consequently, the late baseline carries no information about \(w_0\), although generator noise can leave covariance-shaped pseudo-signal.
Specifically, when \(\sigma_{\mathrm{gen}}=0\), \(\mu_\infty=\delta_0\), where $\delta_0$ is the Dirac delta function at zero. The zero-start baseline is then pure screening noise with covariance \(\Sigma\): if \(G\sim N(0,\Sigma)\),
\(
    \Pi_{\Sigma,D,g}^{\mathrm{gen}}(S_0)=\Pp\{S_0\subset T_D(|G|)\}
\),
where \(T_D(|G|)\) is the set of indices of the \(D\) largest coordinates of \(|G|\).
Corollary~\ref{cor:isotropic-noiseless-random-selection} further gives the cleanest endpoint when \(\Sigma=I_p\). 

\begin{corollary}[Isotropic noiseless random-selection endpoint]
\label{cor:isotropic-noiseless-random-selection}
Suppose \(\Sigma=I_p\), and \(\sigma_{\mathrm{gen}}=0\).
Then, for fixed \(N,r\) and $p$,
\[
    \Pp\{S_0\subset\widehat S_g(D)\}
    \to
    q_{p,D,s}
    :=
    \frac{\binom{p-s}{D-s}}{\binom pD}
    =
    \frac{\binom Ds}{\binom ps},
    ~~\text{as}~~
    g\to\infty.
\]
\end{corollary}

Corollary \ref{cor:isotropic-noiseless-random-selection} shows that if the marginal scores contain no support information and \(\Sigma=I_p\), their ordering is exchangeable, so the selected set is a uniformly random size-\(D\) subset of \([p]\). Among the \(\binom pD\) such subsets, exactly \(\binom{p-s}{D-s}\) contain \(S_0\). The random-selection baseline is therefore \(q_{p,D,s}\):
\[
    q_{p,D,s}
    =
    \prod_{\ell=0}^{s-1}\frac{D-\ell}{p-\ell}
    \le
    \left(\frac Dp\right)^s
\]
when \(D\le p\). Thus \(q_{p,D,s}\to0\) in the usual sparse screening regime \(D/p\to0\) with fixed \(s\), and more generally whenever the screening fraction stays below one and \(s\) grows.


We next turn to a missing unknown-support, minimax result.
Denote $\mathcal S_s=\{S\subset[p]: |S|=s\}$.
Let \(\mathcal W\) be any class of \(s\)-sparse coefficient vectors such that
\(\{\operatorname{supp}(w):w\in\mathcal W\}=\mathcal S_s\) and
\(\sup_{w\in\mathcal W}\|w\|_2\le R\). For \(w\in\mathcal W\), let
\(\Pp_w\) denote the law of the generation-\(g\) screening experiment
\((Z,Y_{g,w}^{\mathrm{scr}})\) initialized at \(w\). Let \(U\sim\operatorname{Unif}(0,1)\)
be independent of every such experiment. A possibly randomized selector is a
measurable map from \((Z,Y,U)\) to a subset of \([p]\) of size at most
\(D\). We suppress \(U\) and write its output as
\(\mathcal A(Z,Y_{g,w}^{\mathrm{scr}})\); deterministic selectors simply ignore \(U\), and all
probabilities below include its randomness. Recall
\(q_{p,D,s}=\binom Ds/\binom ps\) and set
\(\varepsilon_g(R)=\sqrt{N\Lambda}\,\rho_\Sigma^{g/2}R/
\{2\sigma_{\mathrm{sis}}\}\), and define
\[
    \mathfrak R_g
    =
    \inf_{\mathcal A:\,|\mathcal A|\le D}
    \sup_{w\in\mathcal W}
    \Pp_w\{\operatorname{supp}(w)\nsubseteq
    \mathcal A(Z,Y_{g,w}^{\mathrm{scr}})\}.
\]

\begin{theorem}[Late-generation impossibility for screening from recursive labels]
\label{thm:late-impossibility}
For every \(g\ge0\),
\[
    1-q_{p,D,s}-\varepsilon_g(R)
    \le
    \mathfrak R_g
    \le
    1-q_{p,D,s}.
\]
Consequently, for fixed \(N,r\) and $p$ and coefficient class,
\(\mathfrak R_g\to1-q_{p,D,s}\), as \(g\to\infty\). 
\end{theorem}

Theorem~\ref{thm:late-impossibility} strengthens the type of conclusion relative to Theorem~\ref{thm:random-selection-limit}.
The latter compares a fixed screening procedure under the $\theta_0$-initialized and zero-start
experiments; the former optimizes over all size-\(D\) selectors and takes the
worst case over a coefficient class containing every \(s\)-support. Averaging
success probabilities over these supports converts the experiment comparison
into an absolute minimax bound. Here \(q_{p,D,s}\) is combinatorial, rather than
an isotropic-exchangeability consequence: a size-\(D\) output contains only
\(\binom Ds\) of the \(\binom ps\) possible \(s\)-supports.

At finite \(g\), \(\varepsilon_g(R)\) bounds how far residual information about
the initialization can reduce the minimax failure probability below the
random-selection level. For a fixed setup and coefficient class, this advantage
vanishes as \(g\to\infty\), and the minimax failure probability converges to
\(1-q_{p,D,s}\). 
This is a
worst-case conclusion: it does not claim that every support is equally
difficult. 

\subsection{Fixed-Generation Screening and Realized Margins}
\label{subsec:fixed-generation-realized-margins}

The preceding results characterize late-generation information loss and its
minimax consequence. We now return to a fixed generation \(g\) and specialize
to SIS, for which three errors matter around the exact center \(m_g\): generator
fluctuation, labeler-size tails, and empirical score concentration. Define \(v_{\Sigma,\mathrm{gen}}=\sigma_{\mathrm{gen}}^2 r/\{\lambda_*(p-r-1)\}\) and \(b_g=\rho_\Sigma^g\|w_0\|_2^2+v_{\Sigma,\mathrm{gen}}(1-\rho_\Sigma^g)/(1-\rho_\Sigma)\). For \(\delta\in(0,1)\), set \(B_g(\delta)=\Lambda b_g/\delta\).
For \(x>0\), define
\begin{align}
    \psi_N(x,B)
    =
    2p\exp\left[-cN\min\left\{
    \frac{x^2}{C(B+\sigma_{\mathrm{sis}}^2)},
    \frac{x}{C\sqrt{B+\sigma_{\mathrm{sis}}^2}}
    \right\}\right].
    \label{eq:sis-concentration-function}
\end{align}
The constants \(c,C>0\) in \(\psi_N\) are fixed once and for all, with
\(c\) sufficiently small and \(C\) sufficiently large. They are chosen
conservatively enough that the same \(\psi_N(x,B)\) bounds the Bernstein tails
when the deviation threshold is \(x\), \(x/2\), or \(x/4\), as required below.
Let \(C_g=\E\{(\theta_g-\E\theta_g)(\theta_g-\E\theta_g)^\top\}\) and \(V_g^\Sigma=\operatorname{tr}(\Sigma C_g\Sigma)\).

\begin{theorem}[Finite-sample SIS under recursive noisy generation]
\label{thm:unconditional-sis}
Assume that
\(p>r+1\).
For every \(\delta\in(0,1)\), if \(\tau_g^{\mathrm{cen}}(D)>0\), then
\[
\begin{aligned}
    \Pp\{S_0\subset\widehat S_g(D)\}
    \ge
    1
    &-
    \underbrace{16V_g^\Sigma/\{\tau_g^{\mathrm{cen}}(D)\}^2}_{\text{generator fluctuation}}
    -
    \underbrace{\delta}_{\text{labeler-norm tail}}  
    -
    \underbrace{\psi_N\left(\tau_g^{\mathrm{cen}}(D)/2,B_g(\delta)\right)}_{\text{screening-sample concentration}} .
\end{aligned}
\]
If \(\tau_g^{\mathrm{cen}}(D)<0\), then
\[
\begin{aligned}
    \Pp\{S_0\subset\widehat S_g(D)\}
    \le
    &
    \underbrace{16V_g^\Sigma/\{\tau_g^{\mathrm{cen}}(D)\}^2}_{\text{generator fluctuation}}
    +
    \underbrace{\delta}_{\text{labeler-norm tail}} 
    +
    \underbrace{\psi_N\left(|\tau_g^{\mathrm{cen}}(D)|/2,B_g(\delta)\right)}_{\text{screening-sample concentration}} .
\end{aligned}
\]
\end{theorem}

Theorem~\ref{thm:unconditional-sis} is a fixed-generation resolution bound. It converts the centered margin \(\tau_g^{\mathrm{cen}}(D)\) into finite-sample success and failure probabilities by separating three effects. When \(\tau_g^{\mathrm{cen}}(D)>0\), SIS succeeds unless the realized recursive target \(\Sigma\theta_g\) moves too far from its center \(m_g\), the labeler-size event \(\theta_g^\top\Sigma\theta_g\le B_g(\delta)\) fails, or the empirical marginal scores cannot resolve the remaining gap. When \(\tau_g^{\mathrm{cen}}(D)<0\), the centered target already ranks the variables incorrectly; sure screening can then occur only through generator fluctuation, the labeler-size tail, or finite-sample score noise.
Generation \(g\) enters all three components. The center \(m_g\) determines the sign and magnitude of \(\tau_g^{\mathrm{cen}}(D)\), and the covariance \(C_g\) of the recursive labeler controls the generator-fluctuation term. The radius \(B_g(\delta)=\Lambda b_g/\delta\) includes the accumulated generator-noise floor \(v_{\Sigma,\mathrm{gen}}(1-\rho_\Sigma^g)/(1-\rho_\Sigma)\), and the same radius enters the screening-concentration term. Later generations can therefore hurt screening through margin decay, generator fluctuation, and a larger concentration price for screening from a random labeler.

\begin{corollary}[Large screening-sample limit at fixed generation]
\label{cor:fixed-generation-large-N}
Fix $g,p$ and $r$,
and write \(\widehat S_{g,N}(D)\) when the fresh screening sample has size \(N\). If \(\Pp\{\tau(\Sigma\theta_g,D)=0\}=0\), then 
\[
    \Pp\{S_0\subset\widehat S_{g,N}(D)\}
    \longrightarrow
    \Pp\{\tau(\Sigma\theta_g,D)>0\},~~\text{as }~~N\to\infty.
\]
\end{corollary}

Corollary~\ref{cor:fixed-generation-large-N} takes \(N\to\infty\) with \(g\) fixed. A larger fresh screening sample removes empirical score noise, but it leaves the realized target \(\Sigma\theta_g\) random. The limiting sure-screening probability is therefore the probability that this realized target ranks all active variables inside the top \(D\). In contrast, clean-label SIS behaves differently: with deterministic target and \(\tau(\Sigma w_0,D)>0\), sure-screening probability tends to one. Recursive labels break that implication. Furthermore, the centered margin \(\tau_g^{\mathrm{cen}}(D)\) describes the deterministic center, but the large-\(N\) limit depends on the realized margin \(\tau(\Sigma\theta_g,D)\). Thus even a positive centered margin may leave positive probability of a negative realized margin. 

\begin{remark}[Centered and realized margins]
The distinction between the centered and realized margins also clarifies how the realized margin can differ systematically from its center. The map \(v\mapsto\tau(v,D)\) is nonlinear: its active side is a minimum over \(S_0\), while its inactive side is a high order statistic over \(p-s\) competitors. A coordinatewise heuristic is
\[
    \tau(\Sigma\theta_g,D)
    \approx
    \tau_g^{\mathrm{cen}}(D)
    -
    \text{active minimum penalty}
    -
    \text{inactive extreme-value penalty}.
\]
The active penalty comes from the minimum: one downward active fluctuation can reduce the weakest active score. The inactive penalty comes from the order statistic: after absolute values, many inactive coordinates compete, and their upper tail is amplified, typically at a logarithmic scale. These two effects make the realized margin \(\tau(\Sigma\theta_g,D)\) often lie below the centered margin \(\tau_g^{\mathrm{cen}}(D)\). At the same time, the centered margin \(\tau_g^{\mathrm{cen}}(D)\) may shrink with \(g\), which makes the empirical marginal scores have less resolution; the Supplementary Material illustrates this behavior with Gaussian simulations.
\end{remark}


\section{How Screening Collapse Compounds Downstream Model Collapse}
\label{sec:prediction}

Ordinary model-collapse analyses ask how recursive synthetic labels change the fitted predictor. Screening collapse adds an upstream obstruction: by the time the final model is fit, preprocessing has already chosen the variables it may use. We separate this restriction from the later refitting error. For a selected set \(S\), define
\[
    O_\Sigma(S)
    =
    \min_{\operatorname{supp}(b)\subset S}
    (b-w_0)^\top\Sigma(b-w_0).
\]
Thus \(O_\Sigma(S)\) is the best clean prediction risk achievable by any downstream estimator supported on \(S\). The \(\Sigma\)-weight appears because prediction error is measured under \(X\sim N(0,\Sigma)\):
\[
    \E\left[\{X^\top(b-w_0)\}^2\right]
    =
    (b-w_0)^\top\Sigma(b-w_0).
\]
The term \(O_\Sigma(S)\) isolates the loss caused by screening. It is zero whenever \(S_0\subset S\); but if an omitted active variable cannot be reconstructed from the retained covariates, no second-stage estimator supported on \(S\) can remove this loss. 
If \(\Sigma_{SS}\) is nonsingular, the Schur complement gives
\begin{equation}\label{eq:omitted-support-schur}
    O_\Sigma(S)
    =
    w_{0,S^c}^\top
    \left(\Sigma_{S^cS^c}-\Sigma_{S^cS}\Sigma_{SS}^{-1}\Sigma_{SS^c}\right)
    w_{0,S^c}.
\end{equation}
For \(\Sigma=I_p\), this reduces to \(\sum_{j\in S_0\setminus S}w_{0,j}^2\). For correlated designs, the Schur-complement term can be smaller when retained covariates explain omitted active covariates.

The next proposition gives an exact decomposition of the downstream covariance risk into the irreducible omitted-support loss and the additional error incurred relative to the best predictor supported on \(S_g\). Let \(S_g=\widehat S_g(D)\). 
Define \(b_g^\star\in\mathbb R^p\) by
\[
    b_{g,S_g}^\star
    =
    w_{0,S_g}+\Sigma_{S_gS_g}^{-1}\Sigma_{S_gS_g^c}w_{0,S_g^c},
    \qquad
    b_{g,S_g^c}^\star=0.
\]
This vector is the best clean linear predictor on the generation-\(g\) selected set. The following standard projection identity is recorded in a form adapted to the generation-dependent selected set \(S_g\) \citep{berk2013valid,garciarasines2023splitting}.

\begin{proposition}[Covariance-risk decomposition after screening]
\label{prop:post-sis-risk}
For any vector \(\widehat b_g\) supported on \(S_g\),
\[
    (\widehat b_g-w_0)^\top\Sigma(\widehat b_g-w_0)
    =
    \underbrace{O_\Sigma(S_g)}_{\text{omitted-support loss}}
    +
    \underbrace{(\widehat b_{g,S_g}-b_{g,S_g}^\star)^\top\Sigma_{S_gS_g}(\widehat b_{g,S_g}-b_{g,S_g}^\star)}_{\text{refitting, shrinkage, and target error}}.
\]
\end{proposition}

Proposition~\ref{prop:post-sis-risk} makes the downstream effect modular. Once \(S_g\) is fixed, screening has already determined the first term. The second term is the error of the downstream estimator inside the \(S_g\)-restricted problem. If the second stage is ridge regression, this term is the ridge test error in dimension \(|S_g|\), with covariance \(\Sigma_{S_gS_g}\) and oracle restricted target \(b_{g,S_g}^\star\). Classical high-dimensional ridge theory then applies to the selected subproblem; for example, \citet[Proposition~4.2]{dohmatob2024regression} give the clean ridge risk as bias plus variance under Gaussian design. If the post-screening training labels are recursive synthetic labels, their Theorem~4.3 adds the usual model-collapse terms, schematically
\[
    \text{ridge risk on }S_g
    \approx
    \widetilde{\operatorname{Bias}}_{S_g}
    +
    \operatorname{Var}_{S_g}
    +~
    \text{recursive-label noise term},
\]
with ridge degrees of freedom and covariance quantities computed after selection. Screening collapse and ordinary ridge model collapse are additive at this decomposition level: screening creates \(O_\Sigma(S_g)\) by restricting the feature set, and recursive synthetic labels can also increase the refitting term on the surviving variables. 

The finite-sample screening bounds given in Theorem \ref{thm:unconditional-sis} translate directly into bounds for \(O_\Sigma(S_g)\). Let \(\|w_0\|_\Sigma^2=w_0^\top\Sigma w_0\). For \(\delta\in(0,1)\) and \(\tau_g^{\mathrm{cen}}(D)\ne0\), write
\[
    \mathcal E_g^{\mathrm{scr}}(\delta)
    =
    16V_g^\Sigma/\{\tau_g^{\mathrm{cen}}(D)\}^2
    +
    \delta
    +
    \psi_N\left(|\tau_g^{\mathrm{cen}}(D)|/2,B_g(\delta)\right).
\]

\begin{corollary}[Early-generation omitted-support bounds]
\label{cor:early-omitted-support}
If \(\tau_g^{\mathrm{cen}}(D)>0\), then
\[
    \E O_\Sigma(S_g)
    \le
    \|w_0\|_\Sigma^2\,
    \{1\wedge \mathcal E_g^{\mathrm{scr}}(\delta)\}.
\]
If \(\tau_g^{\mathrm{cen}}(D)<0\), 
and
\(\beta_{\min}=\min_{j\in S_0}|w_{0,j}|\), then
\[
    \E O_\Sigma(S_g)
    \ge
    \lambda_*\beta_{\min}^2\,[1-\mathcal E_g^{\mathrm{scr}}(\delta)]_+ .
\]
\end{corollary}

Corollary~\ref{cor:early-omitted-support} controls the screening term at a fixed generation. With a positive centered margin, the expected omitted-support loss is bounded by the same three errors that control sure screening: generator fluctuation, the labeler-norm tail, and screening-sample concentration. With a negative centered margin, the logic reverses: screening misses at least one active variable with high probability unless random fluctuation overturns the negative gap, and that miss forces positive support loss.

To complement this fixed-generation bound, the next result transfers the late-generation comparison in Theorem~\ref{thm:random-selection-limit} to expected omitted-support loss.

\begin{proposition}[Large-generation omitted-support endpoint]
\label{prop:post-sis-population}
Under the assumptions of Theorem~\ref{thm:random-selection-limit},
\begin{align}\label{eq:post-sis-population}
    \left|
    \E O_\Sigma(S_g)-
    \E O_\Sigma(\widetilde S_g)
    \right|
    \le
    \|w_0\|_\Sigma^2
    \frac{\sqrt{N\Lambda}}{2\sigma_{\mathrm{sis}}}
    \rho_\Sigma^{g/2}\|w_0\|_2,
\end{align}
where 
\(\widetilde S_g\) is the selected set under the zero-start noisy-generator baseline. Moreover, for fixed \(N\), let \(\widetilde S_\infty(D)\) be the stationary zero-start selected set from Theorem~\ref{thm:random-selection-limit}, and define
\(
    L_{\Sigma,D,\infty}^{\mathrm{gen}}
    =
    \E O_\Sigma(\widetilde S_\infty(D))
\).
Then,
\[
    \E O_\Sigma(S_g)
    \to
    L_{\Sigma,D,\infty}^{\mathrm{gen}},~~\text{as}~~g\to\infty.
\]
Specifically, if \(\Sigma=I_p\) and \(\sigma_{\mathrm{gen}}=0\), then
\[
    \E O_I(S_g)\to \left(1-\frac Dp\right)\|w_0\|_2^2,~~\text{as}~~g\to\infty.
\]
\end{proposition}

Proposition~\ref{prop:post-sis-population} 
shows that for fixed \(N\), the zero-start selected-set law converges, so \(\E O_\Sigma(S_g)\) converges to \(L_{\Sigma,D,\infty}^{\mathrm{gen}}\). 
In general covariance this limit is implicit, because both the stationary law \(\mu_\infty\) and the omitted-support functional \(O_\Sigma(S)\), represented by the Schur-complement formula \eqref{eq:omitted-support-schur}, depend on \(\Sigma\). In the noiseless identity case, symmetry makes the zero-start selected set uniformly random over all size-\(D\) subsets; each active coordinate is omitted with probability \(1-D/p\), giving \((1-D/p)\|w_0\|_2^2\). If $D/p\to 0$, this is the worst-case limit: the expected omitted-support loss approaches the full signal energy.

The following corollary turns the late support-loss limit into a lower bound for any downstream estimator supported on \(S_g\). 
\begin{corollary}[Irreducible downstream loss from late screening collapse]
\label{cor:irreducible-screening-loss}
Let 
\(\widehat b_g\) be any post-screening estimator supported on \(S_g\). Under the assumptions of Theorem~\ref{thm:random-selection-limit}, 
\begin{align}\label{eq:irreducible-screening-loss}
    \liminf_{g\to\infty}
    \E\left[
    (\widehat b_g-w_0)^\top\Sigma(\widehat b_g-w_0)
    \right]
    \ge
    L_{\Sigma,D,\infty}^{\mathrm{gen}} .
\end{align}
If \(\Sigma=I_p\) and \(\sigma_{\mathrm{gen}}=0\), this gives
\begin{align}\label{eq:irreducible-screening-loss-identity}
    \liminf_{g\to\infty}
    \E\|\widehat b_g-w_0\|_2^2
    \ge
    \left(1-\frac Dp\right)\|w_0\|_2^2 .
\end{align}
\end{corollary}

The refitting method may be ridge, lasso, or another rule. It still cannot use variables that screening removed, so the first term in Proposition~\ref{prop:post-sis-risk} remains even with ideal refitting on \(S_g\).
These results are estimator-agnostic. Screening collapse contributes the upstream term \(O_\Sigma(S_g)\); shrinkage, finite-sample variation, target mismatch and propagated label noise enter through the refitting term on the selected features. We next investigate how to mitigate the upstream screening loss by retaining some real labels or accumulating data.

\section{Real-Label Retention and Data Accumulation for Screening}
\label{sec:accumulation}

Upstream screening failure matters because omitted active variables create loss before the final estimator is fit. Real information may protect the screening step in at least two ways: real-data retention and data accumulation. 
The two mechanisms answer different questions. 
Fixed-fraction real-data retention asks what protection a fixed clean-label component provides when the screening budget does not grow, and where finite-sample screening noise can still defeat it. Data accumulation asks whether a growing data set can resolve the accumulated margin.

\subsection{Fixed-Fraction Real-Label Retention}

We begin with real-data retention. Under the replacement recursion defined in Section~\ref{sec:setup}, once synthetic labels enter, later screening samples contain no observations labeled directly by \(w_0\). The fixed-fraction model allows each screening observation to use the real labeler with probability \(\alpha\) while keeping the screening sample size \(N\) fixed.
Let \(\mathcal F_g\) be the sigma-field generated by the recursive generator up to generation \(g\). Fix a real-label retention fraction \(\alpha\in[0,1]\). Conditional on \(\mathcal F_g\), draw
\begin{equation}\label{eq:fixed-retention-labeler}
    B_i\stackrel{\mathrm{iid}}{\sim}\operatorname{Bernoulli}(\alpha),
    \qquad
    U_{i,\alpha,g}^{\mathrm{mix}}
    =
    B_iw_0+(1-B_i)\theta_g,
    \qquad i\in[N],
\end{equation}
independently of the fresh screening design and noise, and observe
\begin{equation}\label{eq:fixed-retention-response}
    Y_i^{\mathrm{mix}}
    =
    Z_i^\top U_{i,\alpha,g}^{\mathrm{mix}}+\xi_i,
    \qquad
    Z_i\sim N(0,\Sigma),
    \qquad
    \xi_i\sim N(0,\sigma_{\mathrm{sis}}^2).
\end{equation}
Let \(\widehat S_{\alpha,g}^{\mathrm{mix}}(D)\) be the SIS set computed from these labels. The conditional mean labeler is
\(
    u_{\alpha,g}
    =
    \E[U_{i,\alpha,g}^{\mathrm{mix}}\mid\mathcal F_g]
    =
    \alpha w_0+(1-\alpha)\theta_g
\),
so the sample-level mixture has centered margin
\[
    t_{\alpha,g}(D)=\tau(m_{\alpha,g},D)~~\text{with}~~m_{\alpha,g}
    =
    \Sigma\{\alpha w_0+(1-\alpha)M_\Sigma^g w_0\}.
\]
Any \(\alpha>0\) keeps a nonzero fraction of labels tied to the original signal. 
We state the late endpoint first, and then return to fixed \(g\).

\begin{proposition}[Large-generation endpoint under fixed-real-fraction retention]
\label{prop:fixed-real-mixture-limit}
Assume \(p>r+1\) and \(\sigma_{\mathrm{sis}}>0\). Then, 
\[
    t_{\alpha,g}(D)
    \to
    \alpha\tau(\Sigma w_0,D),~~\text{as}~g\to\infty.
\]
Let \(\widetilde S_{\alpha,g}^{\mathrm{mix}}(D)\) be the SIS set obtained from \eqref{eq:fixed-retention-labeler}--\eqref{eq:fixed-retention-response}, with \(\theta_g\) replaced by the zero-start process \(\widetilde\theta_g\) from Theorem~\ref{thm:random-selection-limit}. Then
\[
    \left|
    \Pp\{S_0\subset\widehat S_{\alpha,g}^{\mathrm{mix}}(D)\}
    -
    \Pp\{S_0\subset\widetilde S_{\alpha,g}^{\mathrm{mix}}(D)\}
    \right|
    \le
    \frac{\sqrt{N\Lambda(1-\alpha)}}{2\sigma_{\mathrm{sis}}}
    \rho_\Sigma^{g/2}\|w_0\|_2 .
\]
Let \(\widetilde\theta_\infty\sim\mu_\infty\) be the stationary zero-start law from Theorem~\ref{thm:random-selection-limit}, independently of a fresh screening sample, and let \(\widetilde S_{\alpha,\infty}^{\mathrm{mix}}(D)\) be the corresponding mixed SIS set that uses labeler \(w_0\) with probability \(\alpha\) and \(\widetilde\theta_\infty\) with probability \(1-\alpha\), independently across screening observations. Define
$
    \Pi_{\alpha,\Sigma,D,\infty}^{\mathrm{mix}}(S_0)
    =
    \Pp\{S_0\subset\widetilde S_{\alpha,\infty}^{\mathrm{mix}}(D)\}
$.
For fixed \(N,r\) and $p$,
\[
    \Pp\{S_0\subset\widehat S_{\alpha,g}^{\mathrm{mix}}(D)\}
    \to
    \Pi_{\alpha,\Sigma,D,\infty}^{\mathrm{mix}}(S_0),~~\text{as}~~g\to\infty.
\]
\end{proposition}

Proposition~\ref{prop:fixed-real-mixture-limit} says two things. The deterministic center does not vanish when \(\alpha>0\): it approaches \(\alpha\Sigma w_0\), so the limiting centered margin is \(\alpha\tau(\Sigma w_0,D)\). Generator noise leaves this center unchanged, although it changes the stationary mixture baseline through \(\mu_\infty\). At the same time, the actual mixed screener approaches the zero-start mixed endpoint. The comparison error gains a factor \(\sqrt{1-\alpha}\), because only synthetic-labeled observations differ in the coupling used in the Supplementary Material proof. For fixed \(N\), the mixed sure-screening probability converges to \(\Pi_{\alpha,\Sigma,D,\infty}^{\mathrm{mix}}(S_0)\), not necessarily to one.

The endpoint probability is especially transparent when \(\Sigma=I_p\) and \(\sigma_{\mathrm{gen}}=0\). Then the zero-start endpoint labeler is \(0\). With \(B_i\stackrel{\mathrm{iid}}{\sim}\operatorname{Bernoulli}(\alpha)\), \(Z_i\stackrel{\mathrm{iid}}{\sim}N(0,I_p)\), and \(\xi_i\stackrel{\mathrm{iid}}{\sim}N(0,\sigma_{\mathrm{sis}}^2)\), 
the endpoint SIS score is
\(
    \widehat\omega_{\alpha,\infty}^{\mathrm{mix}}
    =
    N^{-1}\sum_{i=1}^N
    B_i Z_iZ_i^\top w_0
    +
    N^{-1}\sum_{i=1}^N Z_i\xi_i
\).
Hence, \(\Pi_{\alpha,I,D,\infty}^{\mathrm{mix}}(S_0)\) is the finite-\(N\) chance that SIS, computed from a thinned clean-signal sample plus screening noise, retains all active variables. 


Fixed-real-fraction retention preserves a positive clean component in the population target, but with \(N\) fixed, screening noise and inactive extremes may still cause failure. Having characterized the large-generation endpoint, we now return to a fixed generation \(g\) and quantify this finite-sample behavior.

\begin{corollary}[Sample-level fixed-real-fraction retention]
\label{cor:fixed-real-mixture}
Assume the setup of Theorem~\ref{thm:unconditional-sis}. For \(\delta\in(0,1)\), define
$B_{\alpha,g}^{\mathrm{mix}}(\delta)
    =
    (w_0^\top\Sigma w_0)\vee B_g(\delta)$.
If \(t_{\alpha,g}(D)>0\), then
\[
\begin{aligned}
    \Pp\{S_0\subset\widehat S_{\alpha,g}^{\mathrm{mix}}(D)\}
    \ge
    1
    &-
    \frac{16(1-\alpha)^2V_g^\Sigma}{t_{\alpha,g}(D)^2}
    -\delta  
    -
    \psi_N\left(\frac{t_{\alpha,g}(D)}2,
    B_{\alpha,g}^{\mathrm{mix}}(\delta)\right).
\end{aligned}
\]
If \(t_{\alpha,g}(D)<0\), then
\[
\begin{aligned}
    \Pp\{S_0\subset\widehat S_{\alpha,g}^{\mathrm{mix}}(D)\}
    \le
    &
    \frac{16(1-\alpha)^2V_g^\Sigma}{t_{\alpha,g}(D)^2}
    +\delta
    +
    \psi_N\left(\frac{|t_{\alpha,g}(D)|}2,
    B_{\alpha,g}^{\mathrm{mix}}(\delta)\right).
\end{aligned}
\]
\end{corollary}

Corollary~\ref{cor:fixed-real-mixture} is the fixed-generation counterpart of Theorem~\ref{thm:unconditional-sis}. 
At fixed \(g\), the same retention fraction affects all three terms: centered margin, generator fluctuation, and screening-sample concentration radius.
The centered target satisfies
\(
    m_{\alpha,g}
    =
    \Sigma w_0
    +
    (1-\alpha)\Sigma(M_\Sigma^g-I_p)w_0
\),
so the margin-perturbation bound in the Supplementary Material gives
\(
    t_{\alpha,g}(D)
    \ge
    \tau(\Sigma w_0,D)
    -
    2(1-\alpha)
    \|\Sigma(M_\Sigma^g-I_p)w_0\|_\infty
\).
Larger \(\alpha\) protects the centered margin and reduces generator fluctuation by \((1-\alpha)^2\). 
However, observation-level mixing introduces additional heterogeneity: individual responses are generated from either \(w_0\) or \(\theta_g\). The concentration bound must therefore control both component labelers through \(B_{\alpha,g}^{\mathrm{mix}}(\delta)\).

\subsection{Data Accumulation and the Rank Transition}
\label{subsec:literal-accumulation}

We now turn to data accumulation, closer to the mechanism analyzed by \citet[Sections~3.1--3.2]{gerstgrasser2024inevitable} but adapted to screening. Let \(\{X_h\}_{h\ge0}\) be independent generator-training designs in \(\mathbb R^{r\times p}\), each with iid \(N(0,\Sigma)\) rows. Independently of all designs, let \(\eta_h\stackrel{\mathrm{iid}}{\sim}N(0,\sigma_{\mathrm{gen}}^2I_r)\). The initial block has real labels
\[
    Y_0=X_0w_0+\eta_0.
\]
For \(h\ge1\), the new synthetic block is generated by the current accumulated generator,
\[
    Y_h=X_h\theta_{h-1}^{\mathrm{acc}}+\eta_h,
\]
and the next generator is fit on all blocks observed so far:
\[
    \theta_g^{\mathrm{acc}}
    =
    X_{0:g}^{\dagger}Y_{0:g},
    \qquad\text{where}\qquad
    X_{0:g}
    =
    \begin{pmatrix}
    X_0\\
    \vdots\\
    X_g
    \end{pmatrix},
    \qquad
    Y_{0:g}
    =
    \begin{pmatrix}
    Y_0\\
    \vdots\\
    Y_g
    \end{pmatrix}.
\]
Equivalently, \(\theta_g^{\mathrm{acc}}\) is the minimum-norm least-squares solution while the accumulated design is underdetermined and the ordinary least-squares solution once it has full column rank. The accumulated generator-training sample size is \(R_g=(g+1)r\). The generator starts in the same high-dimensional regime as replacement when \(R_g<p\), then crosses into a lower-dimensional regime once \(R_g\) is comparable with \(p\). The fixed-fraction model has no such rank transition.

We also accumulate the screening observations. Draw independent blocks
\(Z_h\in\mathbb R^{N\times p}\), each with iid \(N(0,\Sigma)\) rows, and
independently draw noise vectors
\(\xi_h\stackrel{\mathrm{iid}}{\sim}N(0,\sigma_{\mathrm{sis}}^2I_N)\),
\(h\ge0\), where \(\xi_h=(\xi_{h,1},\ldots,\xi_{h,N})^\top\).
The screening designs and noises are mutually independent and are independent
of all generator-training designs and noises. Set
\[
    V_0=Z_0w_0+\xi_0,
    \qquad
    V_h=Z_h\theta_h^{\mathrm{acc}}+\xi_h,\quad h\ge1,
\]
and compute SIS from the pooled score
\[
    \widehat\omega_g^{\mathrm{pool}}
    =
    \frac1{N_g}\sum_{h=0}^g Z_h^\top V_h,
    ~~\text{where}~~
    N_g=(g+1)N .
\]
Let \(\widehat S_g^{\mathrm{pool}}(D)\) be the top-\(D\) SIS set computed from \(\widehat\omega_g^{\mathrm{pool}}\).
Conditional on the accumulated generator path, the population target of this pooled screener is \(\omega_g^{\mathrm{pool}}=\Sigma\bar\theta_g^{\mathrm{pool}}\), where \(\bar\theta_g^{\mathrm{pool}}=(g+1)^{-1}\sum_{h=0}^g\vartheta_h^{\mathrm{acc}}\), \(\vartheta_0^{\mathrm{acc}}=w_0\), and \(\vartheta_h^{\mathrm{acc}}=\theta_h^{\mathrm{acc}}\) for \(h\ge1\).
The corresponding realized pooled margin is \(T_g^{\mathrm{pool}}(D)=\tau(\Sigma\bar\theta_g^{\mathrm{pool}},D)\).
Write \(P_0=X_0^\dagger X_0\). Since \(X_0\) has the same distribution as a one-generation design in Section~\ref{sec:setup}, \(\E P_0=M_\Sigma\).
Define the deterministic pooled reference vector and its margin by
\begin{equation}
    \bar\theta_g^{\mathrm{pool,ref}}
    :=\frac{w_0+gM_\Sigma w_0}{g+1},
    \qquad
    t_g^{\mathrm{pool,ref}}(D)
    :=\tau(\Sigma\bar\theta_g^{\mathrm{pool,ref}},D).
    \label{eq:accumulated-pooled-reference}
\end{equation}
The reference vector is the unconditional mean of the pooled labeler \(\bar\theta_g^{\mathrm{pool}}\) when the mean exists. 
If not, it is a deterministic
benchmark. We first identify the limiting reference margin. 

\begin{proposition}[Limit of the accumulated reference margin]
\label{prop:accumulated-centered-margin-limit}
For fixed \(p,r,\Sigma,w_0,D\), as \(g\to\infty\),
\[
    t_g^{\mathrm{pool,ref}}(D)
    \to
    t_\infty^{\mathrm{pool,ref}}(D)
    :=
    \tau(\Sigma M_\Sigma w_0,D).
\]
\end{proposition}

Proposition~\ref{prop:accumulated-centered-margin-limit} gives a deterministic
comparison with Proposition~\ref{prop:centered-margin-vanishes}. Under
replacement, the centered signal is repeatedly contracted and
\(\tau_g^{\mathrm{cen}}(D)\to0\). Under data accumulation, the reference
vector averages the clean labeler \(w_0\) with \(M_\Sigma w_0\), so its
limiting margin is \(\tau(\Sigma M_\Sigma w_0,D)\). If this limit is positive, the accumulated centered margin is eventually bounded away from zero; if it is negative, the accumulated population ranking is eventually wrong.
The next corollary gives a perturbation condition for a positive limit and records the identity benchmark. 
Let \(\gamma=\min(r,p)/p\), \(\tau_0=\tau(\Sigma w_0,D)\), and \(\Delta_{\Sigma,r}=\|\Sigma(M_\Sigma-\gamma I_p)w_0\|_\infty\).

\begin{corollary}[Positive limiting reference margin]
\label{cor:positive-accumulated-limit}

If \(\tau_0>0\) and \(2\Delta_{\Sigma,r}<\gamma\tau_0\), then
\[
    t_\infty^{\mathrm{pool,ref}}(D)
    \ge
    \gamma\tau_0-2\Delta_{\Sigma,r}
    >
    0 .
\]
Specifically, if \(\Sigma=I_p\), then 
\[
    t_\infty^{\mathrm{pool,ref}}(D)
    =
    \gamma\tau_0>0.
\]
\end{corollary}

For a general covariance, write \(K_{\Sigma,r}=\Sigma M_\Sigma\). The exact positivity condition is
\(
    \min_{j\in S_0}|(K_{\Sigma,r}w_0)_j|
    >
    q_D\bigl(\{|(K_{\Sigma,r}w_0)_k|:k\notin S_0\}\bigr)
\).
The limiting reference margin is positive when \(\Sigma M_\Sigma w_0\) keeps enough active mass and avoids large inactive leakage. Corollary~\ref{cor:positive-accumulated-limit} gives one simple route: the clean SIS margin is positive and \(K_{\Sigma,r}w_0\) stays close, in sup norm, to the scalar shrinkage \(\gamma\Sigma w_0\).
A low-level sufficient condition can be stated directly in terms of
\(K_{\Sigma,r}\). Let \(\beta_{\min}=\min_{j\in S_0}|w_{0,j}|\),
\(\beta_{\max}=\max_{j\in S_0}|w_{0,j}|\),
\(d_{\min}=\min_{j\in S_0}|(K_{\Sigma,r})_{jj}|\),
\(R_{\mathrm{act}}=\max_{j\in S_0}\sum_{\ell\in S_0,\ell\ne j}
|(K_{\Sigma,r})_{j\ell}|\), and
\(R_{\mathrm{inact},D}=q_D(\{\sum_{\ell\in S_0}
|(K_{\Sigma,r})_{k\ell}|:k\notin S_0\})\).
A rowwise margin bound in the Supplementary Material gives
\[
    t_\infty^{\mathrm{pool,ref}}(D)
    \ge
    \beta_{\min}d_{\min}
    -
    \beta_{\max}\{R_{\mathrm{act}}+R_{\mathrm{inact},D}\}.
\]
If
\(
    \beta_{\min}d_{\min}
    >
    \beta_{\max}\{R_{\mathrm{act}}+R_{\mathrm{inact},D}\}
\),
then \(t_\infty^{\mathrm{pool,ref}}(D)>0\). The condition separates retained self-signal on active coordinates, active-active interference, and inactive leakage. It certifies a positive limiting reference margin when the active diagonal terms dominate interference and leakage.

We next control the deviations introduced by generator noise and finite-sample screening.
For noisy accumulated fits, the Supplementary Material establishes the
design-measurable expansion
\(\theta_h^{\mathrm{acc}}=P_0w_0+\sum_{m=0}^h L_{h,m}\eta_m\),
where \(P_0=X_0^\dagger X_0\) and \(L_{h,m}\in\mathbb R^{p\times r}\)
maps the generation-\(m\) noise into its contribution to
\(\theta_h^{\mathrm{acc}}\). The matrices \(L_{h,m}\) are defined recursively
in the Supplementary Material. Set
\(H_{g,m}=\sum_{h=\max(1,m)}^g L_{h,m}\), with empty sums equal to zero.
Let \(\mathbf e_j\in\mathbb R^p\) denote the \(j\)-th standard basis vector, and define
\(\nu_g^2=(g+1)^{-2}\max_{1\le j\le p}\sum_{m=0}^g\|\mathbf e_j^\top\Sigma H_{g,m}\|_2^2\). Suppose that \(\ell_p:=\beta_{\min}d_{\min}-\beta_{\max}\{R_{\mathrm{act}}+R_{\mathrm{inact},D}\}>0\).

\begin{theorem}[Accumulated screening under general covariance]
\label{thm:general-accumulated-screening}
 Along a sequence with \(g\to\infty\), assume
\(\|\Sigma(I_p-M_\Sigma)w_0\|_\infty/\{(g+1)\ell_p\}=o(1)\), \(\|\Sigma(P_0-M_\Sigma)w_0\|_\infty=o_p(\ell_p)\), and \(\sigma_{\mathrm{gen}}\nu_g\sqrt{\log p}=o_p(\ell_p)\).
If deterministic \(M_g\) satisfies \(\Pp\{\max_{0\le h\le g}(\vartheta_h^{\mathrm{acc}})^\top\Sigma\vartheta_h^{\mathrm{acc}}>M_g\}\to0\) and \(((g+1)N/\log p)\min\{\ell_p^2/(M_g+\sigma_{\mathrm{sis}}^2),\,\ell_p/(M_g+\sigma_{\mathrm{sis}}^2)^{1/2}\}\to\infty\),
then
\[
    \Pp\{S_0\subset\widehat S_g^{\mathrm{pool}}(D)\}\to1 .
\]
\end{theorem}

The theorem gives a sufficient condition, not a sharp characterization. It instead makes explicit the two mechanisms accumulation needs for screening success. The realized accumulated target must keep a positive active-inactive margin, \(T_g^{\mathrm{pool}}(D)=\tau(\Sigma\bar\theta_g^{\mathrm{pool}},D)>0\), and the growing pooled sample must resolve that margin, in the sense that \(\|\widehat\omega_g^{\mathrm{pool}}-\Sigma\bar\theta_g^{\mathrm{pool}}\|_\infty=o_p\{T_g^{\mathrm{pool}}(D)\}\).
If the first requirement fails, SIS follows the wrong population ranking even with infinitely many screening observations. If the second fails, high-dimensional screening noise hides a positive margin. The displayed assumptions give one low-level route to the two requirements. 
In the isotropic noiseless case, the assumptions can be more transparent.

\begin{corollary}[Isotropic accumulated screening]
\label{cor:literal-accumulation-isotropic-limit}
Suppose \(\Sigma=I_p\) and \(\sigma_{\mathrm{gen}}=0\). 
If \(\Pp\{\tau(P_0w_0,D)=0\}=0\), then for fixed \(p,r,N\) and \(g\to\infty\),
\[
    \Pp\{S_0\subset\widehat S_g^{\mathrm{pool}}(D)\}
    \to
    \Pp\{\tau(P_0w_0,D)>0\}.
\]
Along a sequence with \(p\to\infty\), suppose 
for constants \(c_\beta,C_\beta>0\), \(c_\beta\le\min_{j\in S_0}|w_{0,j}|\le\max_{j\in S_0}|w_{0,j}|\le C_\beta\).
If \(r/(s\log p)\to\infty\) and \(gN(r/p)^2/(s\log p)\to\infty\), then
\[
    \Pp\{S_0\subset\widehat S_g^{\mathrm{pool}}(D)\}\to1 .
\]
\end{corollary}

This corollary separates two limits. 
With fixed \(p,r,N\) and \(g\to\infty\), the pooled screening sample size \(N_g=(g+1)N\) diverges, so the empirical SIS score converges to its pooled population target. That target, however, converges to \(P_0w_0\), the projection of \(w_0\) onto the row space of the initial real-data design. Because this projection remains random and may not preserve the active-variable ranking, the limiting sure-screening probability is \(\Pp\{\tau(P_0w_0,D)>0\}\), not necessarily one.
The high-dimensional part of Corollary~\ref{cor:literal-accumulation-isotropic-limit} identifies a sufficient screening-resolution scale. In the late-generation identity/noiseless benchmark, the accumulated margin is of order \(r/p\), and \(gN(r/p)^2\gg s\log p\) ensures that the pooled sample resolves this shrunken margin.

\begin{remark}[Retention, accumulation, and downstream model fitting]
Fixed-fraction real-label retention and data accumulation affect downstream fitting first through the selected support. If \(S_{\alpha,g}^{\mathrm{mix}}=\widehat S_{\alpha,g}^{\mathrm{mix}}(D)\) or \(S_g^{\mathrm{pool}}=\widehat S_g^{\mathrm{pool}}(D)\), Proposition~\ref{prop:post-sis-risk} applies with \(S_g\) replaced by the corresponding selected set, so the omitted-support term is \(O_\Sigma(S_{\alpha,g}^{\mathrm{mix}})\) or \(O_\Sigma(S_g^{\mathrm{pool}})\). Fixed-fraction retention keeps the screening sample size fixed but moves the screening target toward \(\Sigma w_0\); by Corollary~\ref{cor:fixed-real-mixture}, this can reduce the probability that active variables are removed before fitting. Data accumulation adds a second protection mechanism: it preserves earlier information in the pooled target and increases the screening sample size \(N_g=(g+1)N\). Hence Theorem~\ref{thm:general-accumulated-screening} and Corollary~\ref{cor:literal-accumulation-isotropic-limit} show that accumulation can avoid screening collapse when the accumulated active-inactive margin stays positive and the pooled sample resolves it.

The downstream model-fitting effect then parallels the contrast between data accumulation and replacement emphasized by \citet[Section~3.2, Theorem~2, and Appendix~E]{gerstgrasser2024inevitable}. In their isotropic ridgeless linear-regression model, with old data retained and the per-iteration sample size larger than the fitted dimension, accumulated recursive test error is uniformly bounded across generations, while replacement grows linearly. In the present screened pipeline, accumulation can therefore avoid both upstream screening collapse and downstream model collapse only when these two requirements hold together: the accumulated screener keeps the active support with high probability, making \(O_\Sigma(S_g^{\mathrm{pool}})\) small or zero, and the subsequent accumulated-data refit satisfies the bounded-error conditions in the selected lower-dimensional problem. If the accumulated population ranking is nonpositive, or if screening noise cannot resolve a positive margin, the omitted-support loss appears before the bounded-error downstream argument can help.
\end{remark}


\section{Real-Data Experiments}
\label{sec:realdata-experiments}

We continue to examine Riboflavin regression explored in Section \ref{sec:screening-collapse}. The screeners in Figure~\ref{fig:realdata-pseudolabel} of Section \ref{sec:screening-collapse} use different empirical
score functions. They share the generic structure of
screening collapse: recursive pseudo-labeling can
move the population score vector away from its clean-label target, the selected feature sets drift away from their clean-label
counterparts and downstream loss increases. We then 
investigate whether real-data retention and data accumulation can mitigate the screening collapse observed under replacement. 
The experiments are not a direct test of the Gaussian linear model, but they do check whether
these mitigation effects persist beyond Gaussian linear SIS.



Figure~\ref{fig:realdata-pseudolabel}
reports {\it pseudo-label agreement}, defined as the correlation between the
recursive and real training responses. 
For each train/test split and screener, let \(S^{\mathrm{clean}}\) be the variables selected
from the real training labels and \(S_g\) those selected under the
generation-\(g\) recursive-data regime. The {\it selected-feature overlap} is
\(|S_g\cap S^{\mathrm{clean}}|/D\). {\it Clean-refit loss} is the increase in held-out
MSE after selecting \(S_g\) but refitting ridge on the real labels, relative to
the clean-screening baseline based on \(S^{\mathrm{clean}}\). {\it Full-pseudo loss}
uses the same baseline but fits ridge on the labels supplied by the current
recursive-data regime, and therefore also includes downstream fitting
degradation. Higher agreement and overlap are better, whereas larger losses are
worse. For accumulation, we use explicit fixed-design stacking in the spirit of \citet[Section~3.1]{gerstgrasser2024inevitable}. At each generation, the pseudo-labeler is refitted on all blocks accumulated so far.

\begin{figure}[t]
    \centering
    \includegraphics[width=0.95\textwidth]{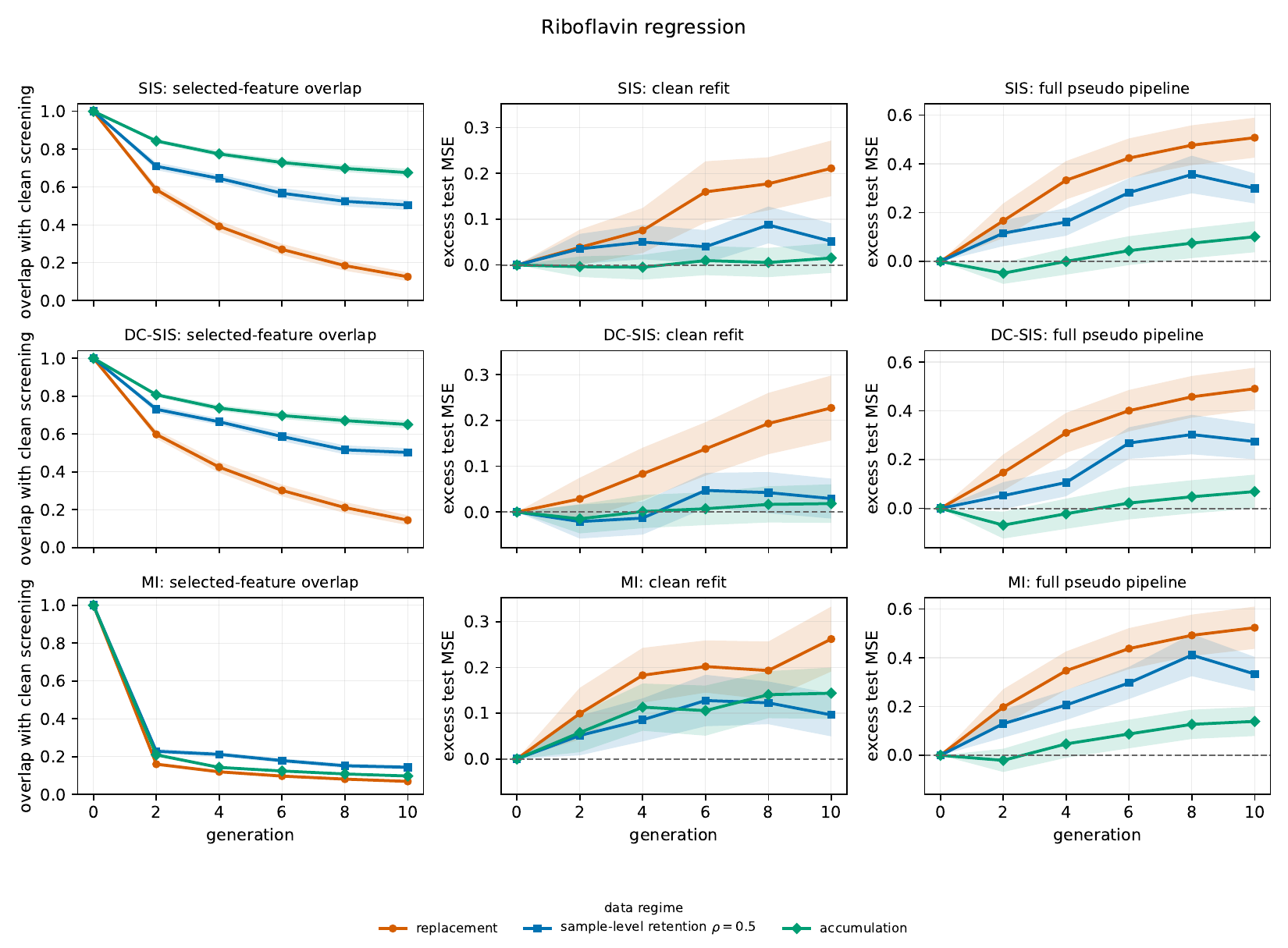}
    \caption{Riboflavin regression under replacement, sample-level retention \((\alpha=0.5)\), and explicit fixed-design accumulation over 100 repeated train/test splits. Rows report SIS, DC-SIS, and mutual-information screening; columns report selected-feature overlap, clean-refit loss, and full-pseudo loss. Curves are means over splits within each screener, with pointwise 95\% normal intervals. }
    \label{fig:realdata-retention-riboflavin}
\end{figure}

Figure~\ref{fig:realdata-retention-riboflavin} reports the selected-feature overlap,
clean-refit loss, and full-pseudo loss under replacement, retention, and accumulation.
The screener-specific trajectories sharpen the comparison. Accumulation most clearly
preserves the clean screening set and controls clean-refit loss for SIS and
DC-SIS; mutual-information screening is less stable, although accumulation
still substantially lowers its full-pseudo loss relative to replacement.
Retention also improves on replacement, generally preserving more of the clean
screening set and reducing both loss measures, although the size of the gain
depends on the screener.
Across all three screeners, accumulation has the lowest full-pseudo loss after
generation zero.

The Supplementary Material reports additional CELLxGENE cell-type
classification experiments using a scTab recursive pseudo-labeler and
distinct cells acquired at each generation.


\section{Discussion and Conclusion}
\label{sec:discussion}

Recursive synthetic data can collapse variable screening before model fitting begins. 
SIS and related marginal screeners rank the current pseudo-label target, not the original sparse real signal. 
Recursive screening collapse is thus a competition between target drift and screening resolution: recursive pseudo-labeling can shrink or reverse the population gap separating active from inactive variables, while high-dimensional empirical fluctuations can overwhelm any remaining positive gap and cause active variables to be discarded.
This upstream failure compounds ordinary model collapse: screening collapse creates the irreducible omitted-support loss, while recursive-label model collapse further degrades fitting on retained variables.
Retention or accumulation can stabilize the marginal gap and reduce omitted-support loss, but neither can guarantee recovery when covariance-induced masking leaves the corresponding margin nonpositive.

The theoretical model is deliberately simple: Gaussian covariates, linear pseudo-labelers, and marginal screening make the failure terms visible. More adaptive generators, iterative screeners, and redundant feature representations may change the severity of screening collapse. 


%


\pdfbookmark[1]{References}{references-bookmark}
\bibliographystyle{plainnat}
\bibliography{references}

\end{document}